\documentclass[letterpaper, 10 pt, conference]{ieeeconf}  

\IEEEoverridecommandlockouts                              

\usepackage{graphicx}
\usepackage{algorithm}
\usepackage{algorithmic}
\usepackage{mathrsfs}
\usepackage{amsmath}
\usepackage{amssymb}
\usepackage{amsfonts}
\usepackage[colorinlistoftodos]{todonotes}
\usepackage{multirow}
\usepackage{subfig}

\usepackage{hyperref}
\usepackage{cite}

\usepackage{hhline}

\title{\LARGE \bf
GAPLE: Generalizable Approaching Policy LEarning for Robotic Object Searching in Indoor Environment
}

\author{Xin Ye$^{1}$, Zhe Lin$^{2}$,  Joon-Young Lee$^{2}$, Jianming Zhang$^{2}$, Shibin Zheng$^{1}$ and Yezhou Yang$^{1}$
                 \thanks{$^{1}$ X. Ye, S. Zheng and Y. Yang are with the Active Perception Group at the School of Computing, Informatics, and Decision Systems Engineering, Arizona State University, Tempe, AZ, USA, Email:
         {\tt \small  \{xinye1, szheng31, yz.yang\}@asu.edu}}
          \thanks{$^{2}$ Z. Lin, J. Lee and J. Zhang are with Adobe Systems, Inc. San Jose, CA, USA, Email:
         {\tt \small  \{zlin, jolee, jianmzha\}@adobe.com}}
 }

\begin{document}

\maketitle
\thispagestyle{empty}
\pagestyle{empty}

\begin{abstract}

We study the problem of learning a generalizable action policy for an intelligent agent to actively approach an object of interest in an indoor environment solely from its visual inputs. While scene-driven or recognition-driven visual navigation has been widely studied, prior efforts suffer severely from the limited generalization capability. In this paper, we first argue the object searching task is environment dependent while the approaching ability is general. To learn a generalizable approaching policy, we present a novel solution dubbed as GAPLE which adopts two channels of visual features: depth and semantic segmentation, as the inputs to the policy learning module. The empirical studies conducted on the House3D dataset as well as on a physical platform in a real world scenario validate our hypothesis,  and we further provide in-depth qualitative analysis. 

\end{abstract}

\section{INTRODUCTION}
Enabling an autonomous robot to search and retrieve a desired object in an arbitrary indoor environment is always both fascinating and extremely challenging, as it would enable a variety of applications that could improve the quality of human life. For example, being able to navigate and localize objects is one of the basic functions that a robot elderly caregiver should be equipped with.
Such technology can also be potentially used to help visually impaired people, thus significantly improving their quality of life. Moreover, self-driving cars with such an object searching capability will be able to approach and pick up their designated customers. Fundamentally, having a robot with vision that finds object is one of the major challenges that remain unsolved. 

With the current surge of deep reinforcement learning \cite{mnih2015human,mnih2013playing,mnih2016asynchronous}, a joint learning method of visual recognition and planning emerges as end-to-end learning \cite{zhu2017target,ye2018active}. Specifically, the robot learns an optimal action policy to reach the goal state by maximizing the reward it receives from the environment. Under the ``robot that finds objects'' setting, the goal state is the location of the target object with a high reward assigned. 
Several recent work have attempted to fulfill the challenge and achieved certain promising results. \cite{zhu2017target} adopted a target-driven deep reinforcement learning model to let robot find a specific visual scene. \cite{ye2018active} also proposed a recognition-guided deep reinforcement learning for robot to find a user-specified target object. Although these deep reinforcement learning models can be trained to navigate a robot to find a target scene or object in an environment, a time-consuming re-training process is needed every time the target or the environment alters. In other words, these systems suffer from an unsatisfiable generalization capability to transfer the previously learned action policy to a brand new target or a novel  environment. Such defect extremely limits the applications of these methods in real-world scenarios as it is impractical to conduct the inefficient training process every single time.
In this paper, we argue that the limitation is deeply rooted in the task itself. While searching an object is indeed environment and object dependent, approaching an object after seen once should be a general capability. The insight could also be explained while observing human beings searching an object in a house. We first need to explore the house to locate the object once. After the object is captured with one sight, we are able to approach the target object with fairly few back and forth explorations. While the exploration policy varies a lot, the optimal approaching policy is indispensable, and provide a critical last step for a successful object search. Thus approaching policy is a much general capability of human beings, and thus in this paper, we focus on the approaching policy learning. We define an approaching task as the robot is initialized  in a state  where the target object can be seen, and the goal is to take the minimal number of steps to approach the target object.

To tackle the challenge, we put forward a novel approach aiming at learning a generalizable approaching policy. We first treat a deep neural network as the policy approximator to map from visual signals to navigation actions, and adopt the deep reinforcement learning paradigm for model training. The trained model is expected to navigate the robot approaching a new target object in a new environment without any extra training effort. To learn an optimal action policy that can lead to a shortest path to approach the target object, previous methods typically attempts to map visual signal to  an optimal action directly, no matter the signal contains clues towards reaching the goal state or not. In such a case, these methods inherently force the policy network to encode the local map information of the environment, which is specific towards a certain scene. Thus, re-training or fine-tuning is needed to update the model parameters while facing a novel target object or a new environment. 

Rather than learning a mapping from each visual signal directly to a navigation action, which has a much higher chance of encoding environment-dependent features, we present a method that first explicitly learns a general feature representations (scene depth and semantic segmentation map) to capture the task-relevant features solely from the visual signal. The representations  serve as the input to the deep reinforcement learning model for training the  action policy. To validate our proposed method's ability to generalize the approaching behavior, empirical experiments are conducted on both simulator (House3D) and in a real-world scenario. We report the experimental results (a sharp increase of the generalization ability over baseline methods) in Section ~\ref{sec:exp}. 

\section{RELATED WORK}
{\bf Target-driven visual navigation.}
Among plenty of methods for target-driven visual navigation, those ones with deep reinforcement learning are most relevant, as we will not provide any human guidance or map related information to the learning system. Recently, \cite{mirowski2016learning} approached the target-driven deep reinforcement learning problem by jointly conducting depth prediction with other classification tasks. \cite{zhu2017target} proposed a target-driven framework to enable a robot to reach an image-specified target scene. \cite{ye2018active} introduced a recognition-guided paradigm for robot to find a target object in indoor environments. Both \cite{das2018embodied} and \cite{gordon2018iqa} aim to let robot navigate in an indoor environment and collect necessary information to answer a question. Although these methods work well in their designed domain, the generalization ability towards new object and new environment is questionable.  More recently, \cite{ammirato2017dataset,mousavian2018visual} attempted to benchmark and study a variety of visual representation combinations for target-driven navigation, showed that using the segmentation and detection mask yields a higher generalization ability.

{\bf Generalization in deep reinforcement learning.}
While generalization ability is a critical evaluation criteria in deep learning, it is less mentioned in the literature of deep reinforcement learning, where most of work focus on improving the training efficiency and the performance of the trained model in certain specific domains \cite{mnih2013playing,mnih2015human,mnih2016asynchronous,jaderberg2016reinforcement,wang2016sample,ghosh2017divide,gu2017deep}. The authors of \cite{dosovitskiy2016learning} proposed to predict the effect of different actions on future measurements, resulting in good generalization ability across environments and targets. However, it is based on the condition of training the model in the complex multi-texture environments. \cite{pathak2017curiosity} adopted the error in predicting the consequences of robot's actions as curiosity to encourage robot to explore the environment more efficiently. Yet it still needs a fine-tuning process when deploying the trained policy in a new environment.  \cite{muller2018driving} addressed the driving policy transfer problem by means of modular design and abstraction, but they only studied the problem w.r.t supervised deep leaning methods, without considering the more challenging generalization issue in deep reinforcement policy learning. 

{\bf Semantic segmentation and depth prediction.}
Semantic segmentation and depth prediction from a single image are two fundamental tasks in computer vision and have been extensively studied. Recently, convolutional neural network and deep learning based methods show dominating performance to both tasks \cite{liu2009beyond,couprie2013indoor,laina2016deeper,chen2018encoder,fu2018deep}. Instead of addressing them separately, \cite{wang2015towards} proposed a unified framework for jointly predicting semantic and depth from a single image. \cite{eigen2015predicting} also adopted a single neural network to do semantic labeling, depth prediction and surface normal estimation. In work \cite{jafari2017analyzing}, the authors analyzed the cross-modality influences between semantic segmentation and depth prediction and then designed a network architecture to balance the cross-modality influences and achieve improved results. Despite the good performance these methods achieved, multi-step training process is still required, that leads to heavy computational load in learning and using these models. In this paper, we adopt a $DeepLabv3+$ \cite{chen2018encoder} based model with which we can perform end-to-end training without a performance loss.

\section{OUR APPROACH}
\label{sec:approach}

\subsection{Overview}

\begin{figure}[ht]
\centering
\includegraphics[width=\columnwidth]{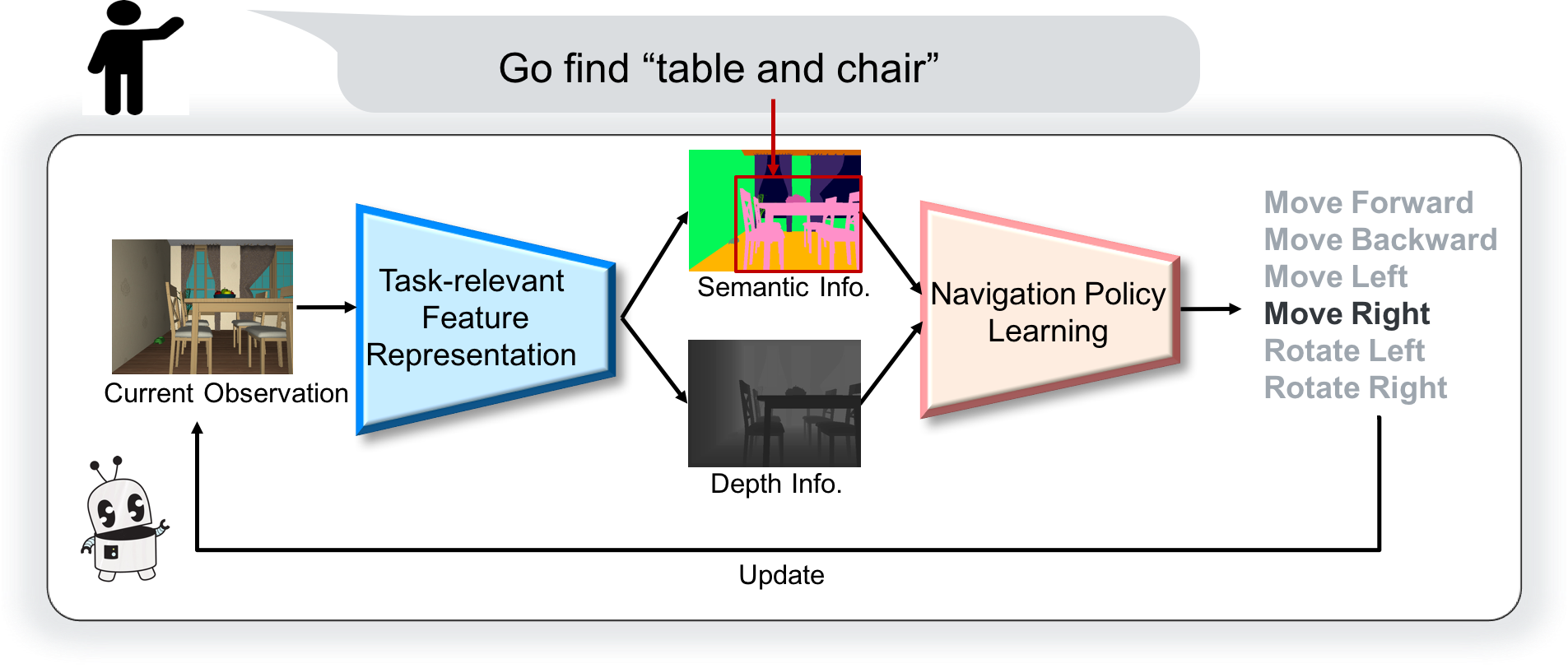}
\caption{An overview of our GAPLE system.}
\label{fig:overview}
\end{figure}

We define the task as learning a generalizable action policy for a robot to approach a user-specified target object with minimal steps. The target object is specified as its semantic category. The robot's on-board camera is the only sensor to capture RGB images, which serve as the robot's observations. The robot starts at the location where the target object can be detected. With the current observation, the robot makes a decision upon which action to take. Afterwards, the robot receives a new observation and it repeats the decision process iteratively until it reaches a close enough location to the target object. Moreover, once the  action policy is trained and deployed, the robot is expected to take reasonable number of steps to approach the user-specified target object as soon as the robot sees it, even the target object is from a new category or the environment changes.

Fig.~\ref{fig:overview} shows an overview of our system. Since the action decision depends on the robot's current observation, the RGB image is the input to the system. Besides, to make the system be flexible to the appearance changes of the target object in the same semantic category, we further include its semantic label as part of the input.

To generalize well across various environments, the feature representations from the input image should be also general across all different environments, or so-called environment-independent. Although the deep neural network is well-suited for extracting task-relevant features \cite{donahue2014decaf}, it tends to capture the environment-dependent features. For example, \cite{zhu2017target} also pointed out that a scene-specific layer is needed to capture the special characteristics like the room layouts. As a result, these models that integrally learns feature representation and navigation policies, can be easily over-fitted towards specific environments. To overcome this challenge, we propose to explicitly learn a more general feature representations. Consider our object approaching task as an example, the robot needs to capture the semantic information from its observation to identify the target object. At the same time, the depth information is crucial for the robot to navigate and avoid collisions. Thus, we adopt a feature representation module that captures both the semantic and the depth information from the input image. We further pipeline the outputs of our feature representation module as the inputs to our proposed navigation policy learning module for action policy training.  The following sections introduce the feature representation module and the navigation policy learning module respectively.

\subsection{Semantic Segmentation and Depth Prediction}
\begin{figure}[ht]
\centering
\includegraphics[width=\columnwidth]{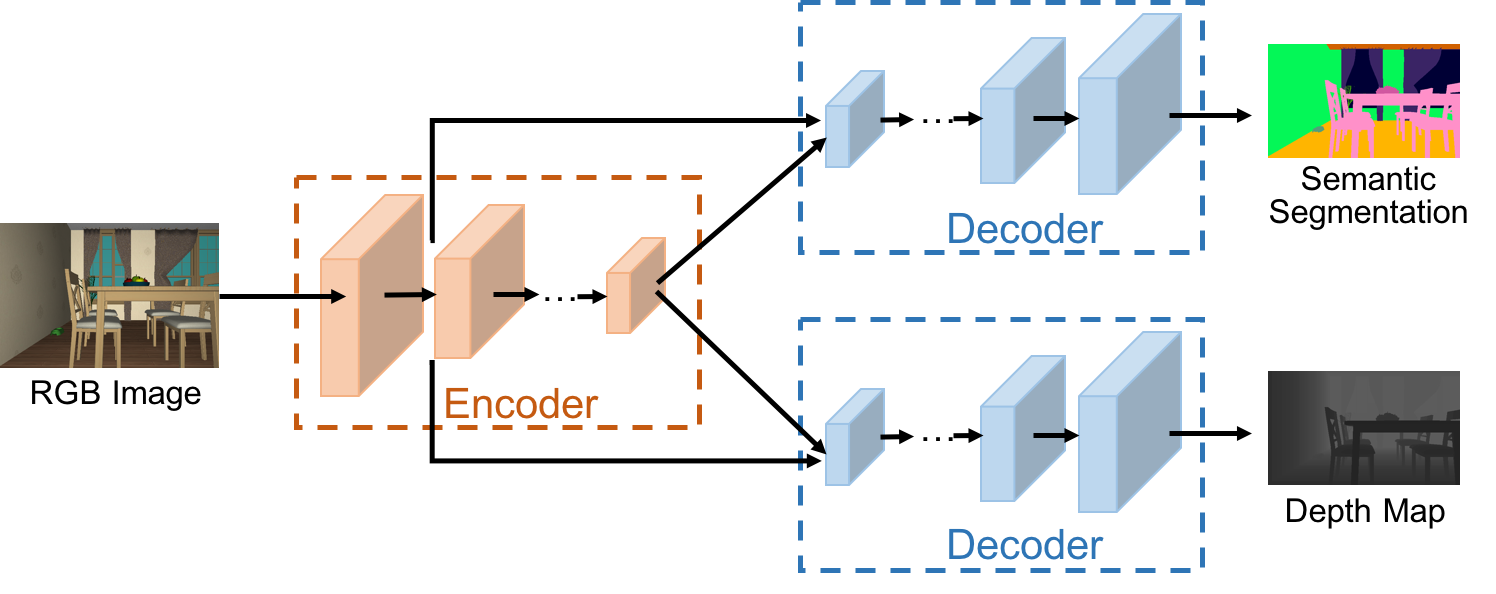}
\caption{An illustration of the adopted model  based on $DeepLabv3+$ \cite{chen2018encoder} to predict semantic segmentation and depth map from a single RGB image.}
\label{fig:segmentation}
\end{figure}

 We adopt the $DeepLabv3+$ \cite{chen2018encoder} based model (as shown in Fig.~\ref{fig:segmentation}) to jointly predict the semantic segmentation and depth map from a single RGB image. $DeepLabv3+$ employs an encoder-decoder structure, where the encoder utilizes the spatial pyramid pooling  to encode multi-scale contextual information. Specifically, it applies multiple filters or pooling operations that with different rates on the feature map computed by other pretrained models, such as ResNet-101 \cite{he2016deep} and Xception \cite{chollet2017xception}. That allows the filters or the pooling operations to be able to consider different field-of-views, so that they can capture rich semantic information. During the decoding process, it gradually up-samples the encoder's output and recovers the spatial information to capture the sharp object boundaries, which leads to better semantic segmentation results. We refer interested readers to \cite{chen2018encoder} for more details.

For generating the depth map at the same time, we spawn another  decoder branch. The motivation is that the depth information and semantic segmentation are correlated. Either one can be used as a guidance to help predicting the other one according to \cite{couprie2013indoor,liu2009beyond}. Thus it is benefiting to jointly predict both of them \cite{wang2015towards,eigen2015predicting}. Here, we adopt the exactly same architecture as the one for semantic segmentation except for the output layer. Rather than a classification layer that outputs labels for each corresponding pixel, we utilize a regression layer instead to predict depth value for each pixel. 
\begin{equation}
\small
\label{eq:loss}
\begin{aligned}
\mathbf{L} = \frac{1}{N}\sum_{i}^{N}{(-\mathbf{p_i}^*log(\mathbf{p_i}))}
+\lambda \frac{1}{N}\sum_{i}^{N}{\Vert d_i-d_i^*\Vert_2^2}
\end{aligned}
\end{equation}

Specifically, our system adopts the Xception-$65$ model \cite{chollet2017xception} pretrained on ImageNet \cite{russakovsky2015imagenet} as initialization. We then define the loss function (Eq.~\ref{eq:loss}) to train our model in an end-to-end manner. The first term is the cross entropy loss for semantic segmentation. $\mathbf{p_i}^*$ is the one-hot encoded ground-truth semantic label for pixel $i$. $\mathbf{p_i}$ is the corresponding predicted probabilities over all possible semantic labels. The second term is the mean-square error for depth prediction, where $d_i^*$ denotes the ground truth depth value for pixel $i$ and $d_i$ represents the corresponding predicted depth value. $N$ denotes the total number of pixels in the image and $\lambda$ denotes a balancing factor. In practice, $\lambda = 0.01$ achieves good performance empirically and we train our model by minimizing the loss function through the stochastic gradient decent (SGD) optimization.

\subsection{Approaching Policy Learning}
\label{sec:drl}
With the semantic segmentation and the depth information computed as the representations of the 
robot's current observation, the robot is expected to make a decision of which action to take to approach the target object. Consider the challenge that the overall {\bf state space} for robot is unknown and each state is of high dimension, we apply the deep reinforcement learning method.

First, we design a deep neural network as an estimator of the policy function. The {\bf policy network} takes both semantic segmentation and depth information as inputs and outputs a probability distribution over all {\bf valid actions} (also known as action policy) . The robot picks a valid action either randomly or follows the distribution predicted by the policy network. After performing the action, the robot receives a {\bf reward signal} as a measurement of how beneficial the performed action is towards the goal. This one-step reward (or likewise the accumulated rewards after taking multiple steps) serves as the weight factor of taking the performed action as the ground truth action for training. We further introduce each part of our setting in details here.

\begin{figure}[ht]
\centering
\includegraphics[width=\columnwidth]{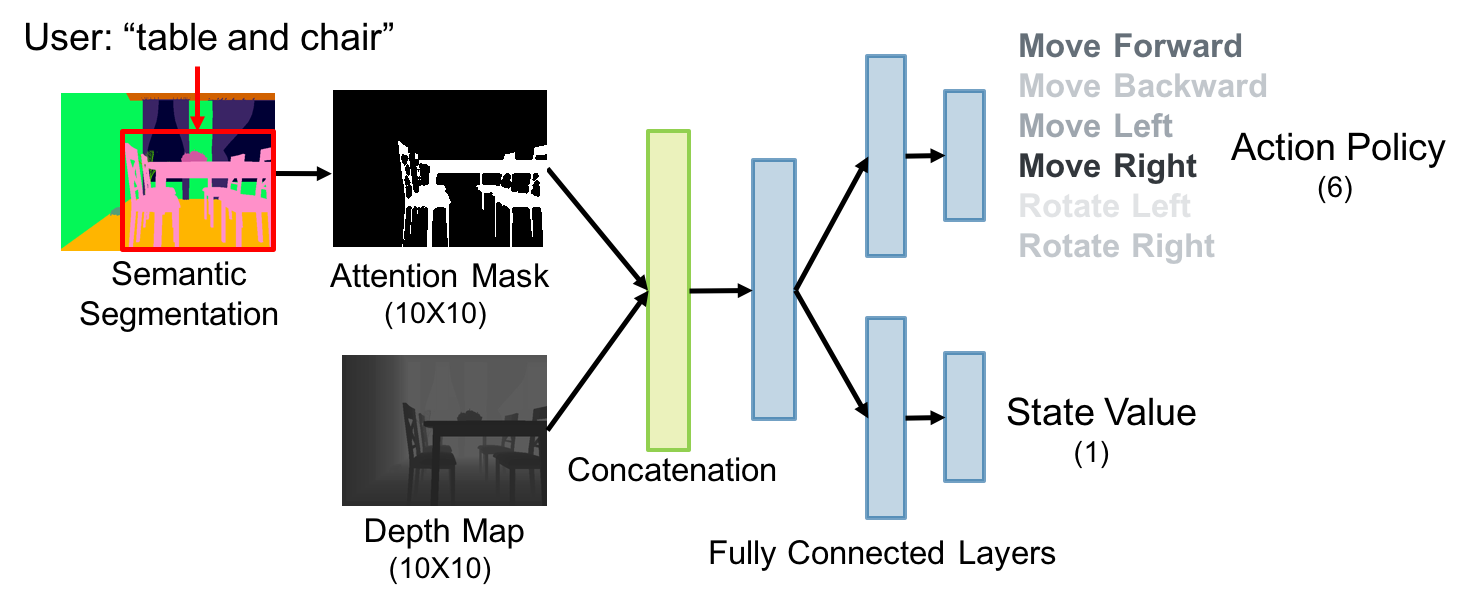}
\caption{The architecture of our deep reinforcement learning model for action policy training.}
\label{fig:drl}
\end{figure}

{\bf State space:} 
Since we assume that the RGB image captured by robot's camera is the only source of information, and both of the robot's position and the target object's location are unknown, the robot's state can only be represented by the captured RGB image as well as the semantic label of the target object. As mentioned before, we represent the RGB image using semantic segmentation and depth map, and the semantic segmentation together with the semantic label of the target object can be further encoded as an attention mask. Afterwards, the attention mask and the depth map together represent the robot's state (see left side of Fig.~\ref{fig:drl}). In addition, the size of the attention field also encodes how close the robot to the target object. Thus, we set a threshold and set the {\bf goal states} as those with an attention field larger than the threshold. In practice, we set it as the size of fifth largest attention field among all ground truth attention masks to yield five goal states. All the possible states form the  state space.

{\bf Action space:} 
To constrain the number of possible states, we only consider discrete actions. Without loss of generality, we consider some basic actions for the navigation purpose, namely ``move forward'', ``backward'', ``left or right with a fixed distance'', ``rotate left or right with a constant angle''. In our experiments, we define a fixed distance as $0.2$ meters and a constant angle as $90$ degrees. 

{\bf Reward function:}
We adopt the reward function designed by \cite{ye2018active} to avoid getting stuck in certain suboptimal states. We define the reward as the size of the attention field if and only if the attention field is larger than all previous ones the robot has observed. Otherwise, the reward is set to be zero. Formally, let $a_t$ be the size of the attention field the robot observes at time step $t$, the reward at this time step $r_t = a_t$ if and only if $a_t>a_{t-1},a_{t-2}...,a_0$, otherwise $r_t = 0$. As a result, the discounted cumulative reward for one episode will be $\gamma^{i_1}a_{i_1}+\gamma^{i_2}a_{i_2}+...+\gamma^{i_t}a_{i_t}$, where $\gamma$ is the discount factor for time penalty and $a_{i_1} < a_{i_2} < ... < a_{i_t} (i_1 < i_2 < ... <i_t)$.

{\bf Policy network:}
Fig.~\ref{fig:drl} illustrates the overall policy learning architecture. The learning module takes the semantic segmentation and depth map as inputs. The semantic segmentation is then used to create an attention mask with the semantic label of the target object. We further resize both the attention mask and the depth map to the size of $10$ by $10$, and then concatenate them into a joint vector before attaching a fully connected layer to generate an embedding fusion. The embedding fusion is then feed into two separate branches, each of which consists of two additional fully connected layers to predict action policy and the state value respectively, where the state value is defined as the expected cumulative reward the robot would receive at the current state. 
 
We follow the training protocol from \cite{zhu2017target}. It trains the model by running multiple threads in parallel and each thread updates the weights of the global shared network asynchronously. However, rather than assigning each thread a specific environment-target pair, we adopt a scheduler with a work stealing manner (\cite{chen2007scheduling}) in order to train all environment-target pairs equally, in case of certain environment-target pairs are much easier to train.

\section{EXPERIMENTS}
\label{sec:exp}

\subsection{Dataset}

\begin{figure}[ht]
\centering
\begin{tabular}{cc}
\subfloat[A sample environment]{\includegraphics[width=0.53\columnwidth]{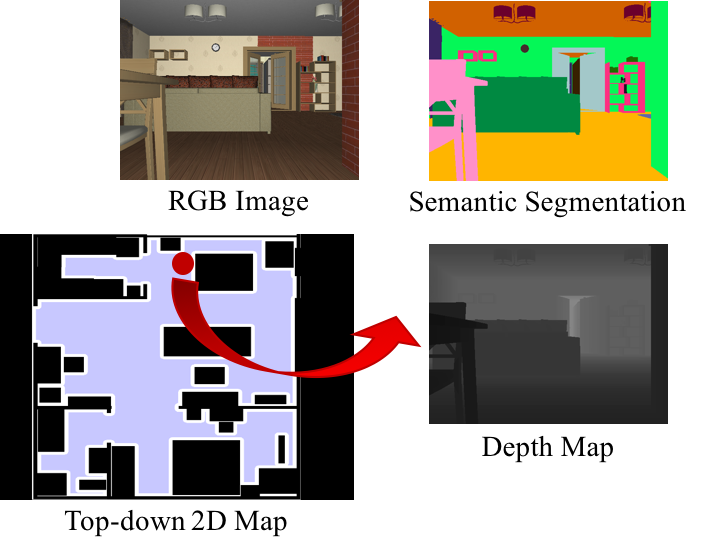}}&
\subfloat[Target object candidates]{\includegraphics[width=0.38\columnwidth]{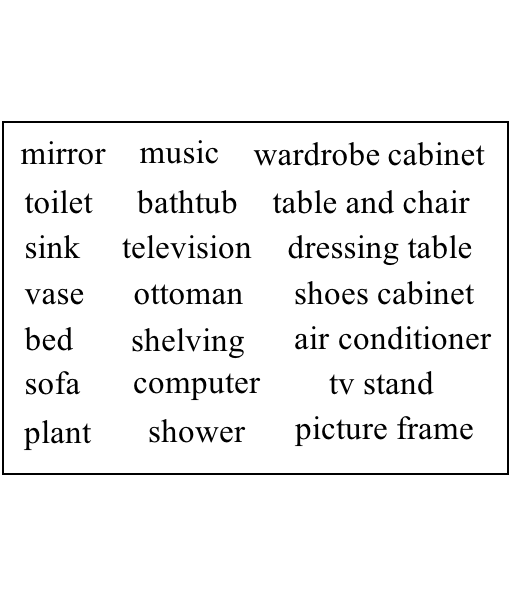}}
\end{tabular}
\caption{A sample environment from House3D and some target object candidates.}
\label{fig:simu_data}
\end{figure}

To train our model and test its generalization ability across different target objects and environments, we need a data efficient platform that has a diverse set of objects and environment types. Here, we adopt the publicly available simulation platform House3D \cite{wu2018building}, a renderer that builds on SUNCG dataset \cite{song2013predicting}. Because House3D consists of rich 3D indoor environments for a virtual robot to interact with. During the interaction, the robot has access to the first-person view RGB images, as well as the corresponding ground truth semantic segmentations and depth maps, which makes it well suited to the feature representation learning task and the approaching policy learning task. Fig.~\ref{fig:simu_data} (a) depicts an example data.

We constrain the robot to perform discrete actions in these virtual environments, i.e. moving $0.2$ meters or rotating $90$ degrees every time. It also discretizes the environment into a set of reachable locations. We select a total of $248$ simulated environments that are suitable for testing. Additionally, to avoid ambiguity, we select the objects that only have one instance in an environment as the target objects for robot to approach. Fig.~\ref{fig:simu_data} (b) lists example target objects used.


\subsection{Semantic Segmentation and Depth Prediction}
In order to train our feature representation module for semantic segmentation and depth prediction, we collect RGB images, as well as their corresponding ground truth captured at all discrete locations from $100$ environments. We further delete the images that has over $80$\% background and randomly sample a total of $55,697$ images for training. For semantic segmentation, $77$ semantic labels are of our interest, with all the remaining ones being classified as ``background''.

\begin{figure*}[ht!]
\centering
\begin{tabular}{ccccccc}
\includegraphics[width=0.35\columnwidth]{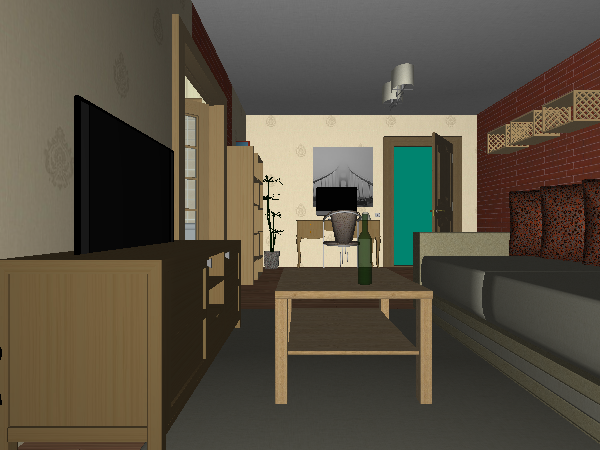}&
\includegraphics[width=0.35\columnwidth]{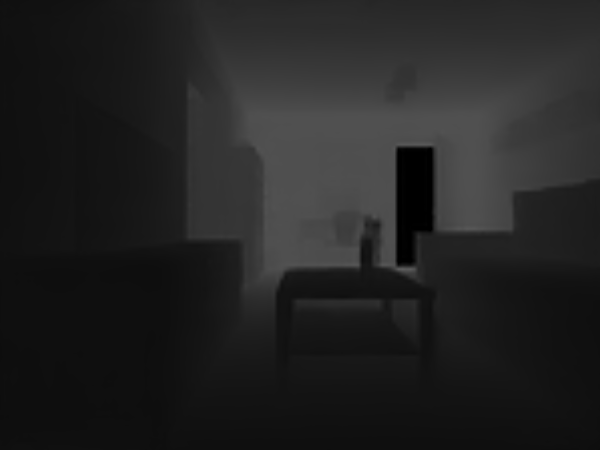}&
\includegraphics[width=0.35\columnwidth]{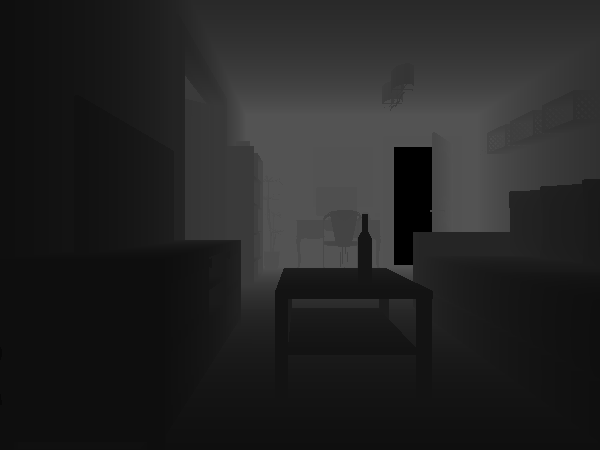}&
\includegraphics[width=0.35\columnwidth]{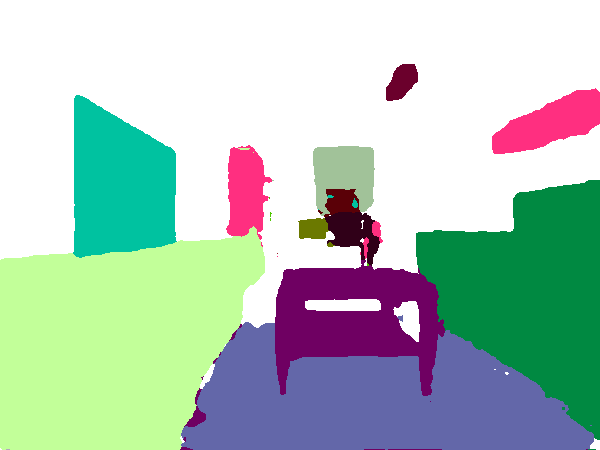}&
\includegraphics[width=0.35\columnwidth]{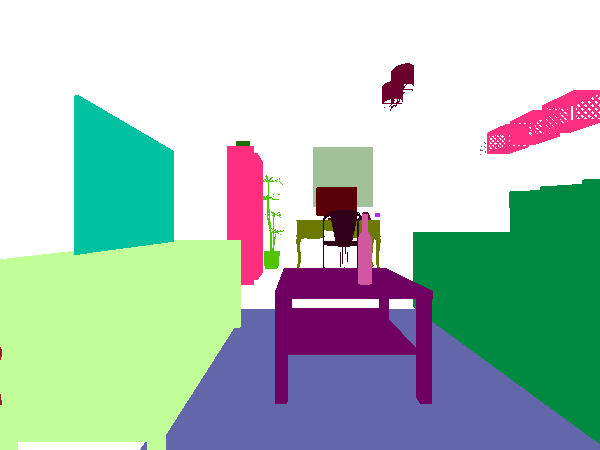} \\
RGB Image & Pred. Depth & GT Depth & Pred. Seg. & GT Seg.
\end{tabular}
\caption{Some qualitative semantic segmentations and depth predictions from our feature representation module.}
\label{fig:seg_depth_results}
\end{figure*}

We take the popularly used metrics, a.k.a. mean Intersection Over Union (mean IOU) and Root Mean Square Error (RMSE) to report the performance of our trained models in doing semantic segmentation and depth prediction respectively. Our model achieves $0.436$ in mean IOU for semantic segmentation on validation dataset, and $0.0003$ normalized RMSE for depth prediction. Fig.~\ref{fig:seg_depth_results} shows several qualitative results.

\subsection{Approaching Policy Learning}
To demonstrate the generalization ability of our proposed method across both target objects and environments, we compare our method with the following baselines and variants. Again, the map of the environment is unknown to all methods, except when calculating the minimal steps that robot needs to take to approach the target object.

\noindent {\bf a)} Random method. At each state, the robot randomly choose an action to perform. Since the map is unknown, the action might yield collision. In that case, the robot will simply stuck in the current state. The random method provides a performance lower-limit, which could calibrate how ``intelligent'' the other trained models are. 

\noindent {\bf b)} Method from \cite{ye2018active}. It takes the output from the $res4f$ layer of ResNet-50 network that pretrained on ImageNet as the feature representation from both the target object and the robot's current observation. The two channels of feature representations, as well as a binary attention mask that is generated using an object recognition module form the inputs to the deep reinforcement learning model.  Here, we first use the ground truth attention mask to remove the influence from the noisy object recognition module. Then we test the method with the attention mask generated from our predicted semantic segmentation. Moreover, we adopt a single scene-specific branch for all the target objects and environments.


\noindent {\bf c)} Our method with ground truth semantic segmentation and depth map. We take the ground truth semantic segmentation and depth map as the inputs to our deep reinforcement learning model as described in Sec.~\ref{sec:drl}, for the purpose of testing the performance upper-limit.

\noindent {\bf d)} Our method with ground truth semantic segmentation and predicted depth map. This method is used to compare with the method b) as they both adopt ground truth semantic info to generate noise-free attention mask.

\noindent {\bf e)} Our proposed method described in Sec.~\ref{sec:approach} that takes only an RGB image and the semantic label of the target object as the inputs, and outputs navigation actions.

We train and evaluate all the methods under two settings for object-wise generalization and environment-wise generalization respectively. {\bf Setting 1):}  training on $6$ different target objects in $1$ environment. The models trained on this setting are then used to evaluate their generalization abilities across target objects. To be specific, during the testing, we use the trained models to approach $5$ unseen target objects in the same environment and report their performances respectively. {\bf Setting 2):} training on a total of $24$ target objects in $4$ different environments ($6$ each). We then test the trained models' performances in approaching another $24$ target objects in $4$ novel and unseen environments. For both settings, the robot always starts at the position where the target object can be detected for both the training and the testing phases.

To conduct fair comparisons, for each testing environment-object pair, we randomly select $100$ starting positions. 
The robot stops either it reaches close enough to the target object (a successful case) or it reaches $1000$ steps (a failure case). We take two metrics to compare these methods, namely the success rate in terms of how many steps (relative to the minimal steps) are taken, and the average steps over the successful cases. Generally speaking, a higher success rate or a smaller number of average steps indicates a better approaching performance. 

\begin{figure*}[ht]
\centering
\begin{tabular}{c}
\includegraphics[width=1\textwidth]{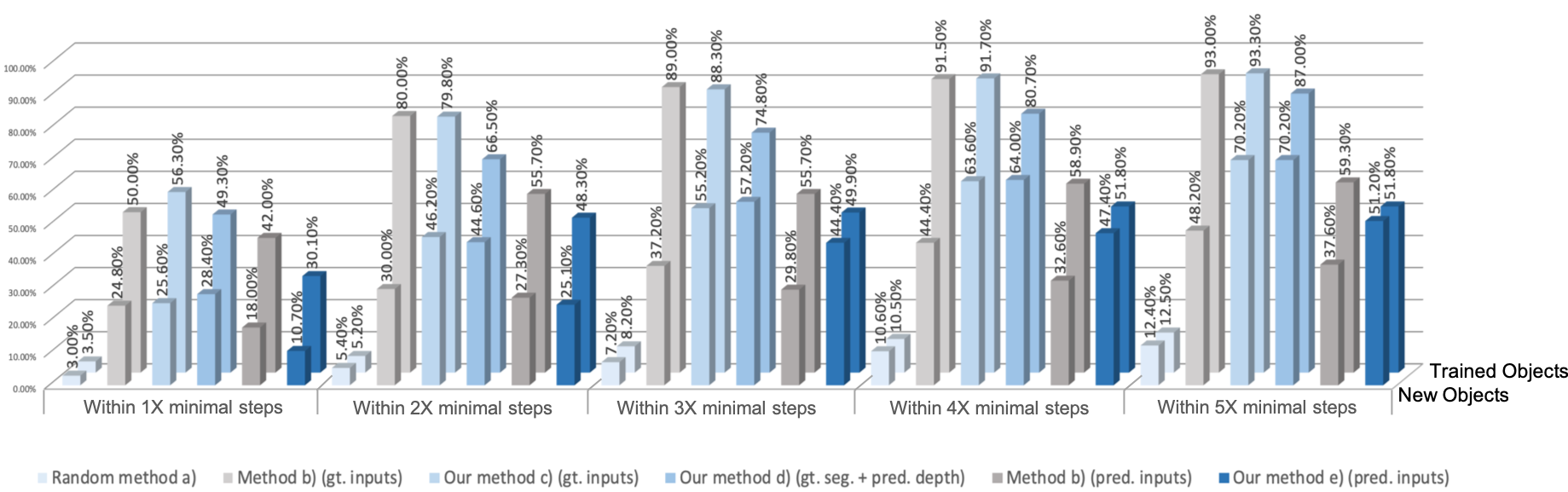} \\
\includegraphics[width=1\textwidth]{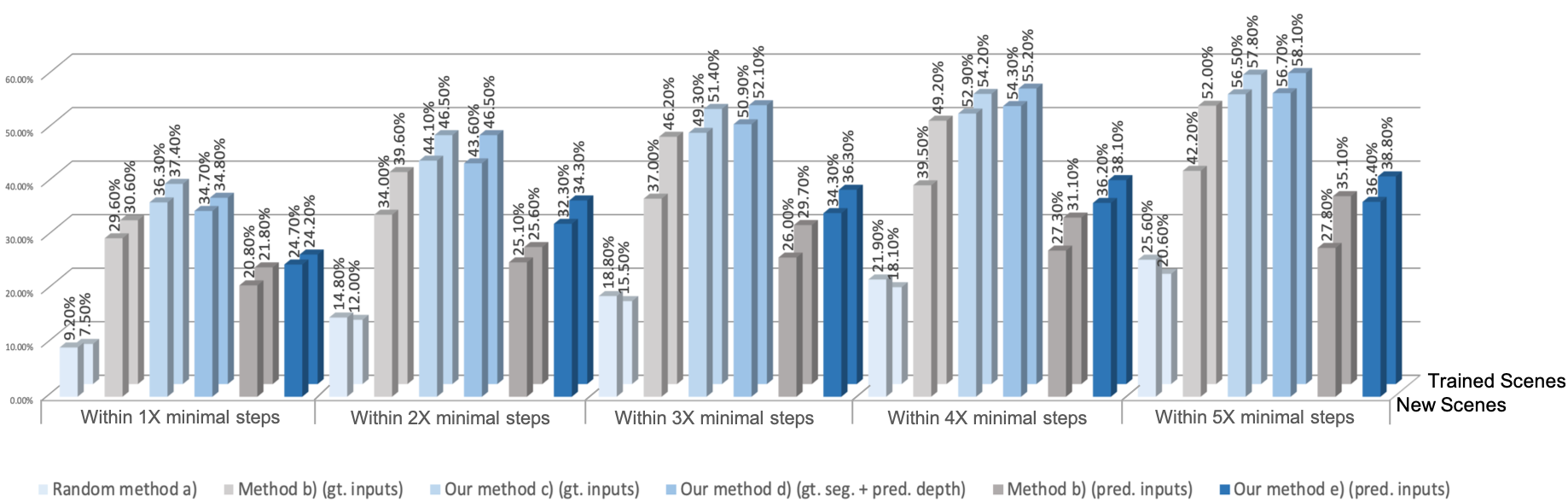} 
\end{tabular}
\caption{Successful approaching rates. Upper: Setting $1$: generalization ability across target objects (on trained objects and on new objects). Lower: Setting $2$: generalization ability across environments (on trained environments and on new environments). 
}
\label{fig:successrate}
\end{figure*}

\begin{table*}[ht]
\caption{The success rate drop from the trained objects to new objects (setting 1: s1), and from the trained environments to new environments (setting 2: s2).}
\label{tbl:success_rate}
\begin{center}
\begin{tabular}{|l|c|c|c|c|c|c|c|c|c|c|}
\hline
\multirow{2}{*}{$\Delta$ success rate} & \multicolumn{2}{|c|}{$1$X minimal steps} & \multicolumn{2}{|c|}{$2$X minimal steps} &  \multicolumn{2}{|c|}{$3$X minimal steps} &  \multicolumn{2}{|c|}{$4$X minimal steps} & \multicolumn{2}{|c|}{$5$X minimal steps} \\
\cline{2-11}
& s$1$ & s$2$ & s$1$ & s$2$ & s$1$ & s$2$ & s$1$ & s$2$ & s$1$ & s$2$ \\
\hline
Random method a) & 0.50\% &-1.70\% & -0.20\% &-2.80\% & 1.00\% & -3.30\% & -0.10\% &-3.80\% &0.10\% &-5.00\% \\
\hhline{|=|=|=|=|=|=|=|=|=|=|=|}
Method b) (gt. inputs) &25.20\% &1.00\% & 50.00\% & 5.60\% & 51.80\% & 9.20\% & 47.10\% & 9.70\% & 44.80\% & 9.80\% \\
\hline 
Our method c) (gt. inputs)&30.70\% &1.10\% & 33.60\% & 2.40\% & 33.10\% & 2.10\% & 28.10\% & 1.30\% & 23.10\% & 1.30\% \\
\hline 
Our method d) (gt. seg. + pred. depth) &20.90\% & 0.10\%& 21.90\% & 2.90\% & 17.60\% & 1.2\% & 16.70\% & 0.90\% & 16.80\% & 1.40\%  \\
\hhline{|=|=|=|=|=|=|=|=|=|=|=|} 
Method b) (pred. inputs) &24.00\%  &1.00\% & 28.40\% & 0.50\% & 25.90\% & 3.70\% & 26.30\% & 3.80\% & 21.70\% & 7.30\% \\
\hline 
Our method e) (pred. inputs) &19.4\% &-0.50\% & 23.20\% & 2.00\% & 5.50\% & 2.00\% & 4.40\%  & 1.90\% & 0.60\% & 2.40\%\\
\hline
\end{tabular}
\end{center}
\end{table*}


\begin{table}[ht]
\caption{Average number of steps taken by all methods on two settings.}
\label{tbl:avgsteps}
\begin{center}
\begin{tabular}{|l|c|c|c|c|}
\hline
\multirow{2}{*}{methods} &\multicolumn{2}{|c|}{setting (1)}& \multicolumn{2}{|c|}{setting (2)}\\
\cline{2-5}
& trained obj. & new obj. & trained env. & new env. \\

\hline
minimal & 3.88 & 3.89 & 2.58 & 2.06 \\
\hline
Random a) & 213.71 & 202.15 & 166.62 & 109.85 \\
\hline
Method b) gt. &5.88 &7.85 &4.34 &3.64 \\
\hline
Method b) pred. &4.70 &5.82 &6.15 &5.94 \\
\hline 
Our method c) &5.85 &13.99 &6.14 &5.14 \\
\hline
Our method d) &8.45 &13.43 & 7.96 &5.96 \\
\hline
Our method e) &10.31 &16.44 & 3.77 &4.22\\
\hline
\end{tabular}
\end{center}
\vspace*{-0.2in}
\end{table}

We trained each of these models with an Nvidia V100 (6 cards with 16g memory each) machine. For setting $1$, each model's training takes about $20$ hours. For setting $2$, the training takes about $40$ hours to converge.

Fig.~\ref{fig:successrate}, Table~\ref{tbl:success_rate} and Table~\ref{tbl:avgsteps} report the achieved success rate, success rate drop from trained objects/environments to new objects/environments, and average steps of all methods on the two settings respectively. From Fig.~\ref{fig:successrate} and Table~\ref{tbl:success_rate}, the results indicate a clear generalization capability improvement of our method d) comparing with the method b) that both take ground truth attention mask, and our method e) with the method b) that both take predicted attention mask. At the same time, the results align well with our expectation that our method with the predicted semantic segmentation (method e)) performs worse than the method d), which also happens for method b). The reason is due to the recognition errors introduced from the predicted semantic segmentation that distracts the robot from approaching the target object. More specifically, the reward generated upon the area of the target object is not consistent due to the noisy detection. Moreover, the area of the robot's attention (focusing on the target object) also needs to encode the goal states. With the noisy predicted semantic labels, the robot has a high likelihood to get stuck while it struggles to identify the correct goal states.

Table~\ref{tbl:avgsteps} reports the average number of steps taken among all successful trails. With the success rate reported in Fig.~\ref{fig:successrate}, it also matches our expectation that the average number of steps from our methods are generally larger than the method b). Here, the larger number of steps means that our methods also succeed in approaching the target object which needs larger number of steps, while the method b) fails these cases and they don't contribute to the average number of steps. 

\subsection{Real World Experiment}
\begin{figure}[ht]
\centering
\includegraphics[width=0.8\columnwidth]{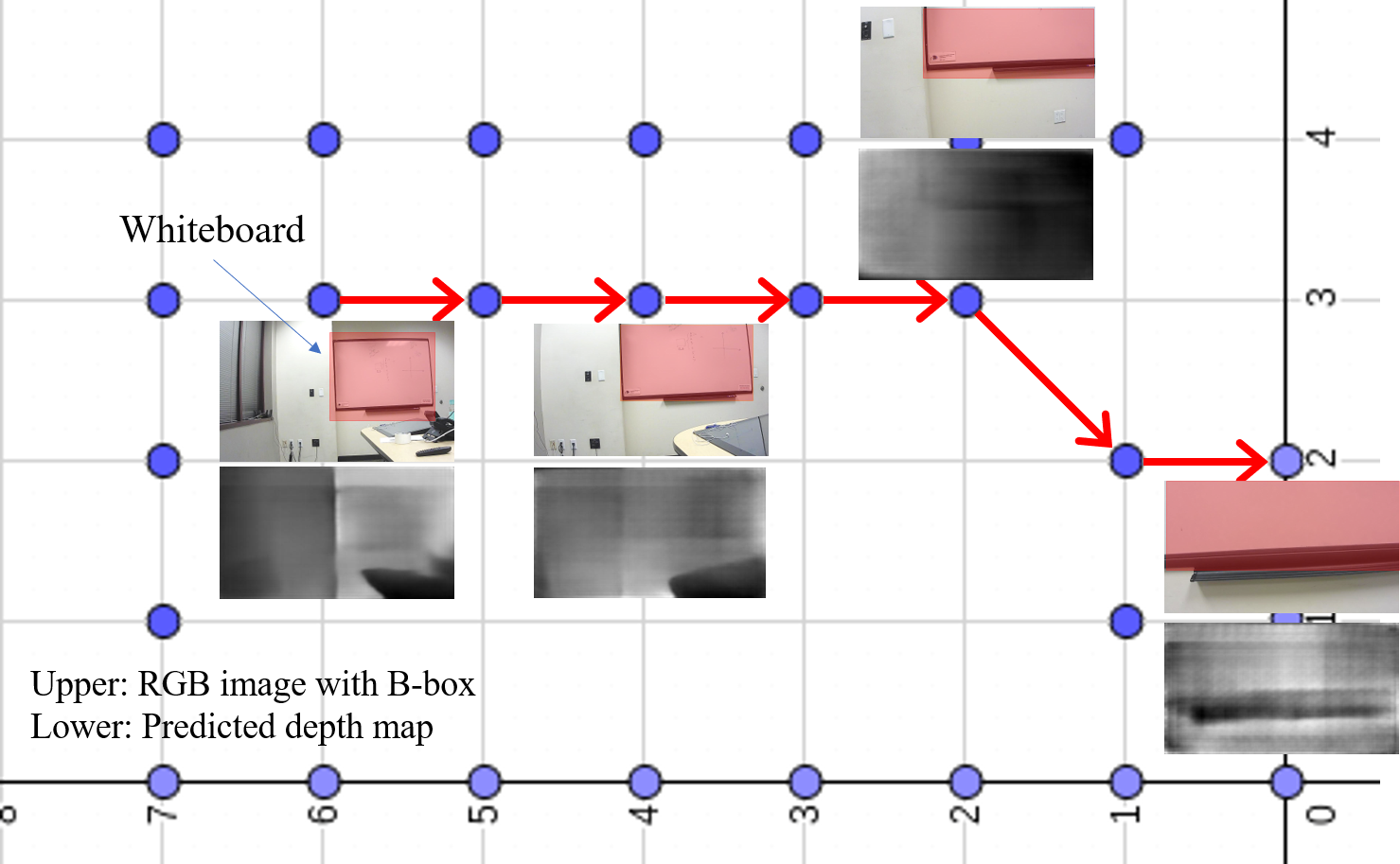}
\caption{An example of the mobile robot approaches the target object ``whiteboard'' using the method (d)). Upper view: RGB input; Lower view: Depth map generated. }
\label{fig:trajectory}
\vspace{-0.2in}
\end{figure}

We adopt our method (method d)) in a real world scenario (on a public dataset from \cite{ye2018active}). Without further fine-tuning the trained model, our trained model can still guide the robot to approach the target object. For this real world experiment, we use the trained model from \cite{laina2016deeper} to predict depth map and the ground truth bounding box to generate the attention mask. Fig.~\ref{fig:trajectory} shows an example of how the robot approaches the target object, which is a  ``whiteboard''.

\subsection{Analysis and Discussion}
To better understand why our proposed method achieves better generalization ability, we further conduct analysis from the feature representation perspective. Generally speaking, the goal of the deep reinforcement learning is to learn a policy model $\pi$ that maps a state $s$ to the most ``beneficial'' action $a$ (or an action distribution from which the most ``beneficial'' action can be drawn with a high probability), i.e. to let $\pi(s) = a$. This most ``beneficial'' action $a$, unlike the ground truth label in a general supervised learning problem, is acquired by the intelligent agent's trail and error interactions with the environment. 

For our object approaching task, the most ``beneficial'' action $a$ essentially depends on the local map between the current location and the goal location. In order to let $\pi(s) = a$, if the input $s$ doesn't provide any map information directly (such as the setting from \cite{ye2018active}), then the model $\pi$ has to capture such information from the input $s$ through learning. To avoid the over-fitting problem, it is necessary to train the model $\pi$ on a large enough and diverse enough training data where the underlying distribution of the relations between the state $s$ and the map information can be captured. Though it is a straightforward, the well-known sample-inefficient issue lingering in the paradigm of deep reinforcement learning makes it fairly impracticable.

In this work, we first adopt a feature representation model $f$ to learn the semantic segmentation and depth map from the input state $s$, then we take the semantic segmentation and the depth map as the inputs to the policy model $\pi$. In other words, we aim to let $\pi(f(s)) = a$. Here, we hypothesize that the depth map as an input to the policy network $\pi$ encodes the local map well already. In such a way, the policy model $\pi$ is not the only source for capturing the local map information well. At the same time, the feature representation model $f$ directly learns the depth map from the state $s$ in a supervised manner, which is much more sample efficient.

\begin{figure}[ht]
\centering
\includegraphics[width=0.8\columnwidth]{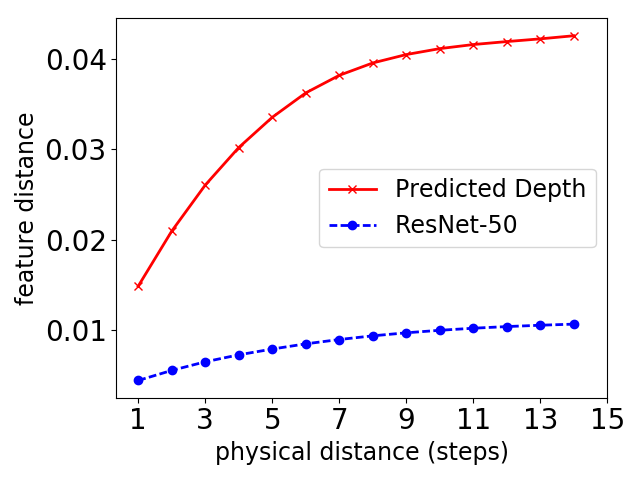}
\caption{Pair-wise feature distances w.r.t. physical distances.}
\label{fig:compare}
\end{figure}

For further validation, we examine the relationship between the distance in physical space and the one in the feature space. For each pair of the locations in an environment, we calculate their Manhattan distance in terms of steps as the physical distance. We adopt $L_1$ distance between the normalized feature maps of the images taken at the two locations with the same orientation as the feature distance. Fig.~\ref{fig:compare} shows the relations between the physical distance and the distance in both depth feature space and ResNet-50 feature space. 

From Fig.~\ref{fig:compare}, it shows that within a small range of physical distances ($1$ to $9$ steps), the distance in depth feature space increases notably along the increment of the physical distance. While the physical distance is out of this range (over $9$ steps), the feature distance shows minor changes. This observation suggests that depth feature captures the differences between different locations within a small region, which aligns well with our assumption. On the other hand, the distance in ResNet-50 feature space grows almost independently w.r.t. the growing of the physical distance. We speculate that this observation provides the actual reason why methods (such as \cite{ye2018active}) fails to generalize well.

\vspace{-0.1in}
\section{CONCLUSION}

This paper presents a novel approaching policy training paradigm dubbed as GAPLE, through explicit depth estimation and semantic segmentation. Empirical studies on the House3D platform and a real physical experiment on a mobile robot validate that the new framework is able to yield a significantly higher generalization capability towards new target objects and novel environments, indicating a promising pathway for future research on achieving generalizable object searching policy on mobile robots.  

\bibliography{references,manipulation,format}
\bibliographystyle{IEEEtran}

\end{document}